\documentclass{article}

\usepackage{PRIMEarxiv}   %改了次文件的Abstract为摘要

\usepackage[utf8]{inputenc} % allow utf-8 input
\usepackage[T1]{fontenc}    % use 8-bit T1 fonts
\usepackage{hyperref}       % hyperlinks
\usepackage{url}            % simple URL typesetting
\usepackage{booktabs}       % professional-quality tables
\usepackage{amsfonts}       % blackboard math symbols
\usepackage{nicefrac}       % compact symbols for 1/2, etc.
\usepackage{microtype}      % microtypography
\usepackage{lipsum}
\usepackage{fancyhdr}       % header
\usepackage{graphicx}       % graphics
\graphicspath{{media/}}     % organize your images and other figures under media/ folder
\usepackage{amsmath}
\usepackage{makecell}
\usepackage{caption}
\usepackage{picins}

\title{A ``Wenlu" Brain System for Multimodal Cognition and Embodied Decision-Making: A Secure New Architecture for Deep Integration of Foundation Models and Domain Knowledge
}

\author{
  {Liang Geng $^{1,2,3}$} \\
  1 College of Mechanical and Electrical Engineering, Shijiazhuang University, Shijiazhuang 050035, China \\
  2 School of Artificial Intelligence, Beijing University of Posts and Telecommunications, Haidian, Beijing 100876, China \\
  3 The Key Laboratory of Agricultural Robotics Intelligent Perception in Shijiazhuang, Shijiazhuang 050035, China \\
  \texttt{Email: liang\_geng@bupt.edu.cn (L.Geng)} \\
}

\begin{document}
\maketitle

\begin{abstract}
With the rapid penetration of artificial intelligence across industries and scenarios, a key challenge in building the next-generation intelligent core lies in effectively integrating the language understanding capabilities of foundation models with domain-specific knowledge bases in complex real-world applications. This paper proposes a multimodal cognition and embodied decision-making brain system, ``Wenlu", designed to enable secure fusion of private knowledge and public models, unified processing of multimodal data such as images and speech, and closed-loop decision-making from cognition to automatic generation of hardware-level code. The system introduces a brain-inspired memory tagging and replay mechanism, seamlessly integrating user-private data, industry-specific knowledge, and general-purpose language models. It provides precise and efficient multimodal services for enterprise decision support, medical analysis, autonomous driving, robotic control, and more. Compared with existing solutions, ``Wenlu" demonstrates significant advantages in multimodal processing, privacy security, end-to-end hardware control code generation, self-learning, and sustainable updates, thus laying a solid foundation for constructing the next-generation intelligent core.
\end{abstract}

% keywords can be removed

\section{Introduction}\label{sec:1}

Amid the continuous evolution of artificial intelligence, large-scale language models (LLMs) driven by deep learning have become a cornerstone in the construction of intelligent systems. However, most mainstream language models today remain predominantly focused on unimodal text-based input-output paradigms, exhibiting limited capabilities in handling multimodal data. Simultaneously, a persistent challenge for both academia and industry lies in how to efficiently integrate domain-specific knowledge with general-purpose models while ensuring the protection of user privacy. To address this gap and advance the deployment of next-generation intelligent applications, this paper proposes an embodied brain system named ``Wenlu", designed to establish a complete closed loop from multimodal cognition to automatic hardware code generation. The system demonstrates a secure and efficient compatibility in scenarios involving private data and domain knowledge, thereby offering a robust foundation for the intelligent integration of large models into real-world tasks.

In recent years, the challenges facing intelligent decision-making and multimodal information processing have primarily arisen from the following aspects: First, different types of data-such as text, images, speech, and sensor readings-differ significantly in dimensionality, feature distributions, and temporal characteristics, making it difficult to efficiently integrate them using a single network architecture or a general-purpose model. Second, deep domain knowledge is often embedded within private or proprietary datasets, requiring more rigorous security and compliance measures. However, foundation models are typically trained on massive open-source data and lack built-in mechanisms for confidentiality and access control, which hinders their direct application in sensitive or regulated fields. Third, many current AI systems remain at the stage of ``perception–cognition" and are unable to automatically translate intelligent decisions into executable processes such as machine control or code generation. This limitation has become a major barrier to the effective deployment of AI in embodied scenarios such as robotics, wearable devices, and autonomous driving.

To effectively address these challenges, the ``Wenlu" system proposes a multi-layered architectural design that tightly integrates a user-private data decision module, an industry-oriented multimodal decision and service module, a hardware control and automated code generation module, and a foundation model fusion unit. The private knowledge unit ensures secure isolation and controllable access to sensitive user information through encrypted sandboxes and labeled data management. The multimodal decision module enables the system to go beyond text processing and perform integrated analysis and reasoning over diverse data sources such as images, speech, and sensor inputs. The hardware control module further bridges the gap between language understanding and physical execution, automatically generating control instructions from high-level cognition or task descriptions to realize real-time feedback and actions on robots or other devices. At the core, the foundation model fusion unit handles comprehensive language understanding and generation, offering robust semantic support to all upper-layer modules, while being deeply coupled with both private knowledge bases and domain-specific datasets.

In addition, drawing inspiration from the biological brain’s mechanisms of memory tagging and replay, the ``Wenlu" system incorporates a brain-inspired memory reinforcement architecture. When executing complex decision-making tasks, the system automatically tags key information and critical reasoning pathways. These tagged memories are subsequently replayed and reinforced during idle or offline phases, enabling continual optimization of multimodal understanding and reasoning capabilities. This brain-like memory mechanism not only enhances the system’s deep cognition and reuse capacity in specific scenarios but also lays a solid foundation for self-learning and continuous iteration.

At the application level, the ``Wenlu" embodied brain system is designed to meet the demands of diverse industries and scenarios, including automated inspection and control in industrial manufacturing, diagnostic assistance in medical imaging, environmental perception in autonomous driving, human–robot interaction in service robotics, and intelligent monitoring in wearable devices. By integrating domain expertise with general semantic understanding, ``Wenlu" offers specialized decision support and automated execution solutions while ensuring data security and privacy compliance. Compared with traditional systems that rely on manual configuration or plug-in invocation, it significantly reduces communication costs and error rates, and achieves self-learning and self-adaptation through continuous iteration.

In summary, the ``Wenlu" system aims to serve as a general-purpose platform for multimodal cognition and embodied decision-making. It inherits the strengths of large language models in textual understanding and generation, while significantly enhancing its capabilities in processing images, audio, sensor data, and other modalities. It enables the secure use of private information and facilitates deep coupling with domain-specific knowledge across various industrial scenarios. Moreover, it translates intelligent decision outcomes into executable hardware control code, achieving a truly integrated closed loop of ``perception–cognition–decision–action". This paper systematically presents the technical framework of the ``Wenlu" embodied brain, focusing on multimodal integration, private information management, automatic code generation, and brain-inspired memory reinforcement. It also provides an in-depth discussion of its potential value across different industries and application domains. Through this architectural design, we aim to offer a novel pathway for transitioning artificial intelligence from academic research to industrial deployment, and to lay a viable technological foundation for the construction of next-generation general intelligent cores.

\section{Theoretical Background and Limitations of Existing Technologies}\label{sec:2}

Over the past decade, artificial intelligence has developed rapidly, evolving from early models based on statistical learning to deep learning, large-scale language models, and multimodal fusion technologies. Especially in the field of natural language processing (NLP), the emergence of general-purpose language models (Large Language Models, LLMs) based on the Transformer architecture has sparked widespread discussion about the potential of ``artificial general intelligence." However, current mainstream technical systems still face numerous challenges in multimodal information integration, private data security, automated code generation, and deep coupling of domain knowledge. To analyze these issues in depth, the following will discuss the existing technical background and limitations in the context of several typical solutions.

\subsection{Challenges in Multimodal Fusion and Cognitive Decision-Making}\label{sec:2.1}

Early applications of artificial intelligence primarily focused on single modalities, such as image recognition, speech recognition, and text generation. However, information in the real world is often multimodal-images, audio, text, and sensor data frequently coexist and interrelate. In this context, multimodal fusion technologies have emerged, aiming to improve perception and decision-making accuracy by establishing unified feature spaces or semantic representations across different modalities.

Although existing multimodal technologies have achieved preliminary success in tasks such as image-text matching and visual question answering, they remain inadequate in deep cognition and complex decision-making, primarily due to the following reasons:
\begin{enumerate}
	\item Difficulty in Feature Alignment and Semantic Projection
	
	The data dimensions and semantic distributions of different modalities vary significantly, necessitating a unified network architecture or feature projection mechanism. Improper alignment among modalities can severely limit the model’s reasoning effectiveness.
	\item Lack of Cross-Modal Long-Term Memory and Reasoning Framework
	
	Most multimodal fusion research focuses on short-term or one-off reasoning, unable to support long-duration, multi-step cognitive processes. For example, industrial inspection or medical diagnosis often requires continuous temporal information and iterative judgment, posing higher demands on multimodal systems.
	\item Insufficient Model Interpretability and Causal Reasoning
	
	Multimodal models typically rely on large-scale training data for pattern matching and have yet to develop an understanding of ``common sense" or ``causal" relationships in the real world. The absence of interpretability mechanisms during industry deployment affects user trust and decision safety.
\end{enumerate}

\subsection{Limitations of Foundation Models and Bottlenecks in Industry Knowledge Integration}\label{sec:2.2}

Since the advent of the Transformer and its improved variants, foundation language models trained on massive public corpora-such as the GPT and BERT families-have made breakthroughs in language understanding and generation. However, directly applying these large models to specific industries or private data scenarios encounters several challenges:
\begin{enumerate}
	\item Lack of Industry-Specific Knowledge
	
	Foundation models are trained on broad datasets covering diverse domains but have limited deep mastery of any specific field. In rigorous professional scenarios such as medical diagnosis or financial analysis, the absence of in-depth domain knowledge significantly reduces the accuracy and reliability of model outputs.
	\item Privacy and Compliance Challenges
	
	Foundation models are often open in nature, lacking strict security isolation in their training data and inference pathways. Mixing users’ private information with public corpora easily leads to data leakage or misuse. Increasingly stringent regulatory requirements on data compliance and user privacy make the secure integration of private data with large models an urgent problem to solve.
	\item Lack of Continual Learning and Memory Mechanisms
	
	Although current large models have vast parameter scales, their ``memory" mainly resides in trained weights, which lack flexibility. Absorbing new knowledge during operation typically involves costly ``fine-tuning" or even ``large-scale retraining," hindering rapid iteration and knowledge accumulation.
	\item ``Breakpoint" from Abstract Understanding to Concrete Execution
	
	Even if a large model can generate accurate textual descriptions, it is difficult to directly convert them into automatically executable hardware commands or robot control scripts. This gap prevents foundation models from fully realizing end-to-end intelligence in embodied scenarios.
\end{enumerate}

\subsection{Traditional Approaches to Robotics and Hardware Control and Their Drawbacks}\label{sec:2.3}

In the field of robotics and hardware control, mainstream development approaches typically require human engineers to manually program action strategies and control logic, translating natural language requirements into low-level commands or API calls. This process has multiple limitations:
\begin{enumerate}
	\item High Costs of Manual Programming and Interface Adaptation
	
	Each robotic platform or wearable device has its specific APIs or SDKs. Porting and adapting these require engineers to spend significant time writing code and debugging, resulting in lengthy development cycles and high susceptibility to errors.
	\item Insufficient Adaptability to Dynamic External Environments

	When the physical environment changes or hardware is updated, substantial program modifications are needed, making it difficult to achieve a dynamic, self-adaptive closed-loop system.
	\item Difficulty Integrating with High-Level Semantic Reasoning
	
	Robotic control often remains limited to sensor data processing and behavior planning, lacking deep understanding of high-level information such as natural language or domain knowledge, and is unable to be unified with multimodal decision-making systems.
\end{enumerate}

\subsection{Compromise Solutions of Plugin-Based / External API Integration and Their Issues}\label{sec:2.4}

To bridge the gap between foundation models and domain-specific applications, some technical practices attempt to augment large models with plugins or external API calls. For example, encapsulating robot APIs or wearable device interfaces as plugins and providing corresponding invocation examples to the language model for extension. While such methods endow foundation models with certain hardware interaction capabilities, they still exhibit the following shortcomings in practical applications:
\begin{enumerate}
	\item Lack of Context Management
	
	Plugin-based approaches inherently rely on temporary API calls and lack a unified memory and reasoning management system, resulting in limited context continuity for large models during multi-turn conversations or cross-modal scenarios.
	\item Complex Cross-Domain Invocation Chains
	
	When a task requires multiple plugins to collaborate simultaneously (e.g., reading sensor data, controlling a robotic arm, and generating diagnostic reports), the plugin invocation chain becomes cumbersome and highly coupled, leading to high maintenance costs.
	\item Unstable Decision Outcomes
	
	Instability in plugin interfaces and invocation timing often causes unexpected system failures or errors. Once an anomaly occurs, the large model itself lacks global awareness of the hardware or external APIs, making self-correction difficult.
\end{enumerate}

\subsection{Knowledge Graph-Based Question Answering and Decision-Making Systems}\label{sec:2.5}

In specific industries such as healthcare, finance, and manufacturing, knowledge graph-based question answering and reasoning systems have also been developed. These systems encapsulate structured expert knowledge into searchable graph nodes and edge relationships to assist inference and decision-making. However, such approaches face several constraints:
\begin{enumerate}
	\item High Construction and Maintenance Costs
	
	Knowledge graphs require dedicated teams for data annotation, updating, and maintenance. Moreover, multimodal features (such as images, audio, and sensor data) are difficult to represent efficiently within the graph.
	\item Lack of Strong Support for Language Generation
	
	Compared with general large models, traditional knowledge graphs mainly focus on retrieval-based question answering or logical reasoning, lacking natural language understanding and generation capabilities, and are insufficient in presenting complex domain data in a readable format.
	\item Inability to Achieve Automated Hardware Output
	
	Knowledge graph-based systems primarily emphasize ``information retrieval" and ``logical reasoning", and remain far from automatically generating robot or device control commands, making it difficult to form an embodied closed-loop system.
\end{enumerate}

\subsection{Overview and Limitations Analysis}\label{sec:2.6}

Combining the above-mentioned typical approaches, it can be summarized that existing technologies exhibit significant limitations in the following aspects:
\begin{enumerate}
	\item Insufficient Multimodal Fusion
	
	Whether relying on general large models or knowledge graphs, the vast majority of existing systems lack the capability to simultaneously process multimodal information such as images, audio, text, and sensor data for deep decision-making, or can only achieve limited functional integration through plugin-based methods.
	\item Lack of Secure and Controllable Private Data Management
	
	Most current large models overlook privacy protection requirements, while traditional industry systems often lack support for open language understanding; efficient and secure integration between the two remains challenging.
	\item ``Break" Between Cognition and Execution
	
	General large models can perform text understanding and generation, but still require manual ``translation" for physical hardware control, resulting in a disconnect between actual decision-making and execution, which impedes self-learning and rapid iteration.
	\item Lack of Continuous Self-learning and Memory Reinforcement Mechanisms
	
	Whether large models or knowledge graphs, updates and fine-tuning usually incur high costs. Most solutions lack brain-inspired memory tagging and replay mechanisms, making it difficult to accumulate new experience and knowledge through continuous interaction and application.
\end{enumerate}

Therefore, to break through the comprehensive barriers of multimodal cognition, private information protection, and embodied decision-making, a novel architecture is needed that combines general language understanding, efficient domain knowledge integration, secure private data protection, and automated hardware code generation capabilities. It is under this background and driven by such demands that the ``Wenlu" embodied brain system emerges, dedicated to constructing an intelligent core for multimodal cognitive decision-making, emphasizing private information security, and capable of directly outputting executable commands, thereby forming a complete closed loop of ``Perception-Cognition-Decision-Execution".

\section{Architecture of the ``Wenlu" System}\label{sec:3}

The ``Wenlu" Embodied Brain system aims to bridge the critical gaps in AI spanning multimodal cognition, private information protection, and hardware control execution, thereby delivering more comprehensive, flexible, and secure intelligent services for practical scenarios. To achieve this goal, ``Wenlu" adopts a multi-layered, modular architecture that includes not only the front-end for external perception and interaction but also a core backend deeply integrating large models with private knowledge bases; simultaneously, it reserves modules for hardware control generation and domain knowledge integration to meet end-to-end application requirements.

The first layer is the User Private Knowledge Understanding and Q \& A Decision Unit, primarily addressing how sensitive user information and domain-specific proprietary data can be securely utilized within large models. This module employs technologies such as secure sandboxes, encrypted storage, and labeled management to ensure strict control over read/write permissions for private data entering the system. When a user query involves sensitive content, the system automatically invokes designated access policies to parse and desensitize the required private data. This unit not only effectively isolates public training data from user private data but also provides highly customized support for subsequent decision-making.

The second layer is the Industry Multimodal Decision and Service Support Unit, whose core function lies in integrating industry-specific multimodal inputs such as images, voice, text, and sensor data with general large models at the semantic level. By converting low-level processing of image recognition, audio analysis, and text understanding into unified feature vectors or embedding representations, this module can comprehensively evaluate the correlations among multisource signals and make deep decisions. In demanding fields like industrial inspection, medical imaging analysis, or autonomous driving monitoring, this multimodal decision unit significantly improves the accuracy and reliability of results and outputs richer interpretable analysis reports.

The third layer is the Robot Control and Hardware Device Code Generation Unit, acting as the bridge from cognition to action within the entire ``Wenlu" architecture. After the multimodal decision unit and private knowledge decision unit collaboratively produce a confirmed execution plan, the robot control unit can automatically generate low-level code scripts from high-level semantic instructions, adapting to various hardware platforms such as ROS2, wearable device APIs, and intelligent robotic arm interfaces. Through this end-to-end automation, the system forms an integrated closed-loop from natural language description and image/sensor fusion analysis to concrete device execution. Unlike traditional manual programming, ``Wenlu" achieves rapid migration upon hardware or environment changes by simply updating the adaptation layer, greatly reducing engineering costs.

Supporting the entire system at the foundation is the Deep Language Model Base and ``General-to-Specialized" Knowledge Fusion Unit. This unit comprises general large models (e.g., DeepSeek) with universal language understanding and generation capabilities, alongside integrated expert domain knowledge bases to enhance Q \& A and reasoning accuracy in vertical fields. Additionally, this unit embeds a brain-like memory tagging and replay mechanism: during each interaction or inference, the system automatically annotates key decisions or critical scenarios and replays and reinforces them during idle or offline periods. This consolidates ``long-term memory" for high-value knowledge and frequent tasks. Such brain-inspired memory consolidation markedly improves system reliability and customization over repeated interactions and offers both theoretical and technical guarantees for integrating large models with private data and domain knowledge.

It is important to emphasize that these units are not isolated components but collaborate efficiently through a unified bus and communication protocol: the private knowledge decision unit shares core model capabilities with the multimodal decision unit and accesses encrypted data based on permission policies; multimodal analysis results are directly passed to the hardware control generation unit to produce executable scripts or instructions; the foundational general model and domain knowledge bases iteratively incorporate distilled content from repeated interactions via memory tagging to provide increasingly accurate support for Q \& A and decision-making.

Through this layered, modular architecture, ``Wenlu" establishes a closed-loop system connecting multimodal perception, semantic decision-making, private data processing, and robot control, offering a powerful and flexible solution for intelligent applications across industries. Within this framework, the system not only performs deep reasoning over multisource data in complex environments but also outputs automated hardware execution scripts while fully safeguarding user sensitive information. Ultimately, ``Wenlu" is poised to become the next-generation general intelligent core, combining cross-industry scalability with personalized domain expertise, thereby providing robust technical support for transitioning AI from laboratories to real-world production and life scenarios.

\section{Core Modules and Implementation Mechanisms}\label{sec:4}

Within the ``Wenlu" Embodied Brain architecture, each module is designed around the principles of multimodal fusion, privacy protection, and an end-to-end decision-making closed loop. To clearly present the technical concepts and internal logic, this section provides an academic and structured exposition of the four core modules, detailing their structures, mechanisms, and inter-module collaboration according to system functionality and implementation flow.

\begin{figure}[htbp]
	\includegraphics[width=\linewidth]{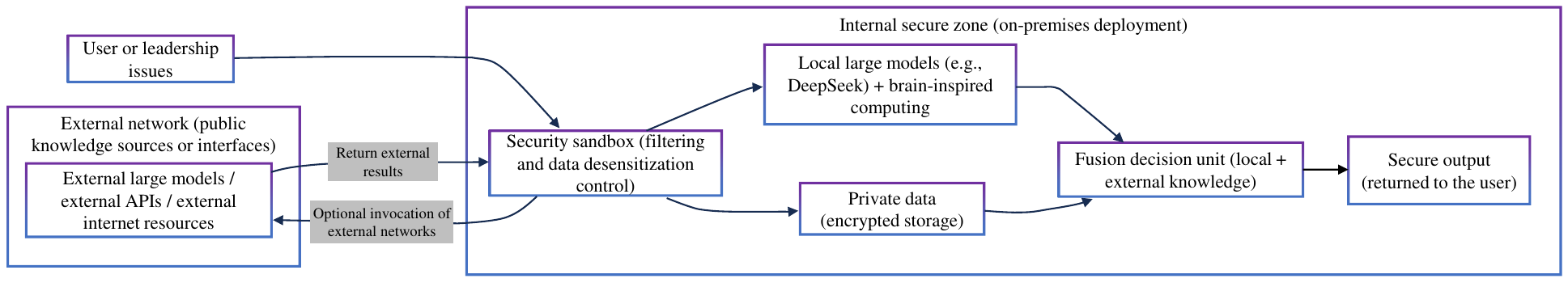}
	\caption{\label{fig01}}
\end{figure}

\begin{figure}[htbp]
	\centering
	\includegraphics[width=0.6\linewidth]{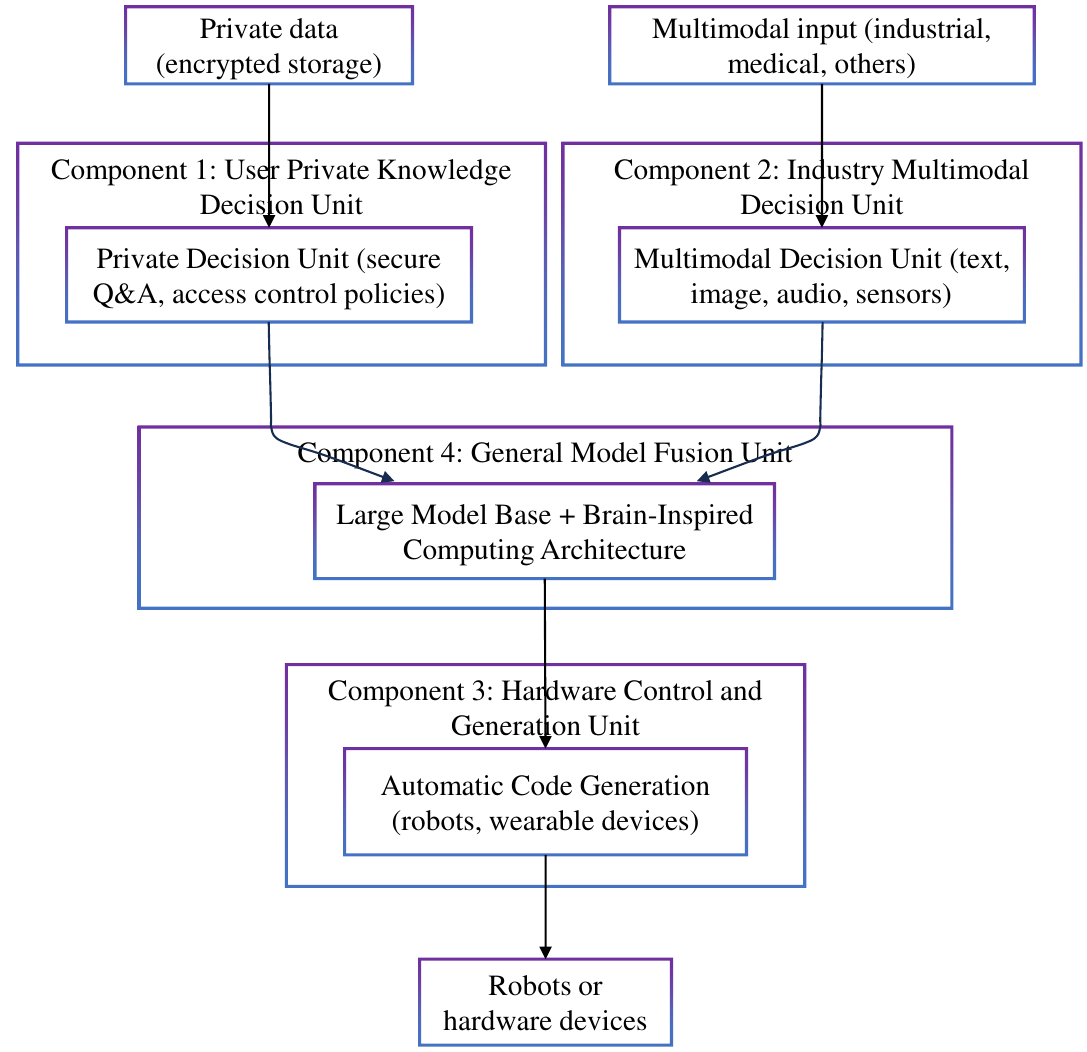}
	\caption{\label{fig02}}
\end{figure}

\subsection{User Private Knowledge Understanding and Q \& A Decision Unit}\label{sec:4.1}

\subsubsection{Module Functionality}\label{sec:4.1.1}

This module focuses on the secure integration of user sensitive information with general large models, ensuring that confidential data is accessed and processed under strict permission controls while satisfying the requirements for efficient inference and customized Q \& A. Its core functionalities include:
\begin{enumerate}
	\item Sandbox-style encrypted storage and labeled management of private data.
	\item Access control and desensitized output for queries or inference tasks involving private data, according to security policies.
	\item Deep coupling of private data with the general language model, enabling precise calls during inference via encrypted indexing mechanisms.
\end{enumerate}

\subsubsection{Technical Implementation Mechanisms}\label{sec:4.1.2}
\begin{enumerate}
	\item Secure Sandbox and Encryption Strategy
	\begin{itemize}
		\item Utilizes symmetric or asymmetric encryption algorithms for storing private data, complemented by Role-Based Access Control (RBAC) for access permissions.
		\item Constructs a ``Private Knowledge Base Index" table, assigning an encryption key and security labels to each confidential piece of information.
	\end{itemize}	
	\item Permission Verification and Desensitized Output
	\begin{itemize}
		\item When a user query involves data flagged by security labels, the permission verification module is triggered first to determine if access is allowed or if desensitization procedures must be executed.
		\item Upon authorization, the private knowledge unit decrypts the original data and converts it into semantic vectors; after merging with the inference results from the general model, a secondary review and sensitive information filtering is performed.
	\end{itemize}	
	\item Collaboration Between Private Knowledge and General Large Model
	\begin{itemize}
		\item Employs encrypted indexing to embed private corpora within the implicit feature space shared by the general model, enabling dynamic retrieval of relevant confidential information during inference.
		\item The final Q \& A or decision output undergoes security policy checks before release to ensure no unauthorized leakage of sensitive information.
	\end{itemize}
\end{enumerate}

\subsection{Industry Multimodal Decision-Making and Service Support Unit}\label{sec:4.2}

\subsubsection{Module Functionality}\label{sec:4.2.1}

The ``Wenlu" multimodal decision-making unit aims to organically integrate various types of industry data (such as images, audio, text, sensor signals) and, supported by industry-specific knowledge bases and general language models, output more refined and interpretable decision results. Its main functions include:
\begin{enumerate}
	\item Unified feature extraction and embedding representation of diverse input modalities.
	\item Context understanding and situational awareness through semantic fusion of multimodal information by large models.
	\item Deep judgment and service output tailored to specific industry scenarios (e.g., medical diagnosis, industrial inspection).
\end{enumerate}

\subsubsection{Technical Implementation Mechanisms}\label{sec:4.2.2}
\begin{enumerate}
	\item Feature Fusion and Semantic Projection
	\begin{itemize}
		\item For data types including images, audio, and sensors, specialized deep neural networks are employed to extract feature vectors; these are then projected and fused with textual inputs within a unified semantic space.
		\item Contextual information across modalities is interconnected using mechanisms such as Multi-Head Attention or Cross-Modal Transformers, generating a comprehensive multimodal representation.
	\end{itemize}
	\item Interaction Between Industry Knowledge Base and General Models
	\begin{itemize}
		\item Based on the fused representation, the system invokes both the general language model and domain-specific knowledge bases for inference. When private data is involved, label matching with the private knowledge unit is also performed.
		\item The process culminates in interpretable industry-specific outputs such as diagnostic reports, fault detection, predictive analyses, or service plans.
	\end{itemize}	
	\item Explainability and Service Interfaces
	\begin{itemize}
		\item To support high-reliability applications in sectors like industry and healthcare, the system integrates explainability components that provide core evidence alongside decision conclusions (e.g., annotated defect regions, semantic key points).
		\item The final results can be delivered via APIs or front-end interfaces for user reference or further interaction.
	\end{itemize}
\end{enumerate}

\subsection{Robot Control and Hardware Code Generation Unit}\label{sec:4.3}

\subsubsection{Module Functionality}\label{sec:4.3.1}

This unit is the key component of the ``Wenlu" system for embodied decision-making. It enables the automatic translation from high-level semantic understanding to executable hardware control instructions, providing direct execution plans for physical platforms such as robots and wearable devices. Its main functions include:
\begin{enumerate}
	\item Generating corresponding executable scripts or API call instructions based on task descriptions output by the general language model or the multimodal decision unit.
	\item Adapting to various robotic operating systems (e.g., ROS2) and other hardware device driver layers, while managing real-time feedback and dynamic environmental adjustments.
	\item Establishing a closed loop from perception to cognition to action, thereby reducing manual programming effort and interface mapping costs.
\end{enumerate}

\subsubsection{Technical Implementation Mechanisms}\label{sec:4.3.2}
\begin{enumerate}
	\item Mapping from High-Level Semantics to Commands
	\begin{itemize}
		\item Leveraging the sequence-to-sequence generation capabilities of language models, this module converts natural language requirements or multimodal reasoning outputs into hardware control languages such as ROS2 node scripts, embedded C++, or Python execution scripts.
		\item During the output stage, the generated code can either be compiled or interpreted to meet the specifications of different operating environments.
	\end{itemize}
	\item Adaptation Layer and Interface Management
	\begin{itemize}
		\item A unified adaptation layer is designed for mainstream robotic frameworks and hardware platforms. This layer translates intermediate instructions generated by the general model into function calls or configuration files that conform to specific APIs, facilitating rapid hardware switching.
		\item When hardware or environmental configurations change, engineers only need to modify the adaptation layer, without the need for large-scale adjustments to high-level logic.
	\end{itemize}
	\item Real-Time Feedback and Closed-Loop Iteration
	\begin{itemize}
		\item Status data or sensor feedback generated during hardware execution can be fed back into the multimodal decision unit for subsequent decision updates, enabling adaptivity and self-learning.
		\item Through continuous iteration, the system can better adapt to complex or dynamic external environments.
	\end{itemize}
\end{enumerate}

\subsection{Deep Language Model Foundation and ``General–Domain-Specific Integration" Knowledge Fusion Module}\label{sec:4.4}

\subsubsection{Module Functionality}\label{sec:4.4.1}

This unit forms the foundational support of the entire ``Wenlu" system, integrating open-source or commercial general-purpose large language models (e.g., DeepSeek) with industry-specific knowledge bases through a ``general–domain-specific" fusion approach. It is responsible not only for cross-domain language understanding and generation, but also for providing a unified semantic foundation for reasoning over multimodal and private data.

\subsubsection{Technical Implementation Mechanisms}\label{sec:4.4.2}
\begin{enumerate}
	\item Pre-trained Model Deployment and Integration with Domain Knowledge Bases
	\begin{itemize}
		\item The system deploys pre-trained general large models on local servers or cloud platforms and incorporates domain expert knowledge bases into the shared semantic space through fine-tuning or incremental training.
		\item In response to varying domain demands, the system can dynamically load corresponding sub-knowledge bases to enable more targeted domain-specific reasoning capabilities.
	\end{itemize}
	\item Neuro-Inspired Memory Tagging and Replay
	\begin{itemize}
		\item Inspired by biological memory tagging and replay mechanisms, the system marks important scenarios or reasoning paths each time it completes a complex decision-making or service task.
		\item During idle or offline periods, it replays and reinforces the tagged information, achieving secondary consolidation of potentially high-value knowledge and gradually improving prediction accuracy and execution efficiency for similar tasks.
	\end{itemize}
	\item Multi-Module Collaboration and Adaptive Updating
	\begin{itemize}
		\item The private knowledge module, multimodal decision module, and hardware control module all communicate bidirectionally with the model foundation to access general language capabilities or update internal knowledge indices.
		\item When new domain-specific data or private information is received, the system uses encrypted indexing and incremental fusion strategies to gradually embed such knowledge into the model architecture, enhancing its deep reasoning and customized service capabilities.
	\end{itemize}
\end{enumerate}

\subsection{Inter-Module Collaboration and Overall Workflow}\label{sec:4.5}

In practical operation, ``Wenlu" forms a closed-loop system by integrating the above four core modules through a unified interface bus and communication protocol. The overall workflow is illustrated as follows:
\begin{enumerate}
	\item Reception of External Multimodal Data and User Requests: The multimodal decision-making unit performs feature extraction or semantic alignment on incoming multimodal data (e.g., images, speech, text, sensor signals) and user queries. If private data is involved, the system invokes the private knowledge unit for secure handling.
	\item Core Language Model Inference and Memory Tagging: The general model fusion unit accesses the foundational language model and domain-specific knowledge bases to conduct deep analysis on multimodal inputs, while simultaneously tagging critical decision paths or scenarios for memory replay and reinforcement.
	\item Secure Processing of Private Information: If the reasoning process or output involves sensitive data, the private knowledge unit is triggered to carry out sandboxed decryption, permission verification, and necessary data desensitization before further usage or output.
	\item Decision Output and Hardware Control: Once a high-level decision is made, the robot control and code generation unit translates it into executable scripts or API calls, which are dispatched to the target hardware, such as robots or embedded systems.
	\item Execution Feedback and Self-Adaptation: Feedback from devices or the external environment is fed back into the multimodal and core language modules, initiating memory replay and reinforcement learning to continuously optimize the overall performance of the ``Wenlu" system.
\end{enumerate}

Through this modular and scalable architecture, ``Wenlu" flexibly supports the full spectrum from software-level question answering to hardware-level execution, achieving truly multimodal cognition and embodied artificial intelligence. Each module complements the others: it ensures the security of private data while also delivering deep decision-making support and automatic execution across diverse industry scenarios.

\section{System Innovations and Technical Advantages}\label{sec:5}

Building upon the deep integration of multimodal fusion, large-scale language model semantic understanding, and secure privacy protection, the ``Wenlu" embodied brain system establishes a relatively complete and closed-loop architecture for AI innovation. The following elaborates on its key innovations and technical advantages from three dimensions: technology, application, and ecosystem.

\subsection{Technological Innovations}\label{sec:5.1}
\begin{enumerate}
	\item Multimodal Brain-Inspired Memory Tagging and Replay Mechanism
	
	Unlike traditional systems that rely solely on the powerful reasoning and generation capabilities of general-purpose large language models (LLMs) in the textual domain, ``Wenlu" introduces a biologically inspired ``memory reinforcement" mechanism:
	\begin{itemize}
		\item It tags key decision pathways from multimodal data-including text, images, audio, and sensor information-and replays them during system idle or offline phases for consolidation.
		\item Through this mechanism, the system accumulates domain-specific experiential knowledge over time, gradually improving its comprehension and reasoning accuracy in complex scenarios.
	\end{itemize}	
	\item Deep Integration and Secure Management of Private Data
	
	To address the prevalent privacy and compliance demands in industry applications, ``Wenlu" incorporates a dedicated private knowledge decision unit and encrypted sandbox architecture:
	\begin{itemize}
		\item Confidential user information is stored and managed in isolation from public general-purpose corpora through encryption and labeling, eliminating the risk of data leakage between private and public datasets.
		\item During inference, any invocation of sensitive data is preceded by strict permission checks and desensitization procedures, ensuring tightly controlled exposure and minimizing security risks.
	\end{itemize}
	\item Embodied Closed-Loop from Cognition to Code Generation
	
	Conventional AI systems often remain at the ``cognition-to-decision" level, requiring manual implementation of control logic to achieve physical execution. In contrast:
	\begin{itemize}
		\item ``Wenlu" integrates a built-in module for robot control and hardware code generation, automatically translating high-level decisions into executable hardware instructions or scripts.
		\item This eliminates the need for repeated manual translation, significantly reducing labor costs and development cycles, while enabling rapid adaptation to changing environments.
	\end{itemize}
	\item Deep Coupling of General-Purpose LLMs with Domain Knowledge
	
	To bridge the gap between industry-specific data and open-domain large models, ``Wenlu" adopts a ``general-to-specific" integration strategy within its general model fusion unit:
	\begin{itemize}
		\item Through lightweight fine-tuning or incremental training, domain-specific knowledge is organically embedded into the semantic space of general-purpose language models.
		\item When applied to vertical industry scenarios, the model can leverage both general and specialized knowledge, producing more precise question answering and complex reasoning outcomes.
	\end{itemize}
\end{enumerate}

\subsection{Technical Advantages and Application Value}\label{sec:5.2}
\begin{enumerate}
	\item Unified Multimodal Processing
	
	The system is capable of handling text, images, audio, and various types of sensor data simultaneously, integrating multimodal inputs into a unified semantic space. This enables higher precision and stronger interpretability in decision-making for scenarios requiring the coordination of diverse information sources, such as industrial inspection, medical diagnosis, and autonomous driving.
	\item Secure and Controllable Private Data Management
	
	``Wenlu" establishes a robust security framework for handling sensitive user information through an independent private knowledge decision module and a tagging-based encryption strategy. This approach strikes a balance between the value of data utilization and compliance risk, while also providing a feasible path for enterprise-level private deployment of large language models.
	\item End-to-End Embodied Code Generation
	
	With its dedicated hardware control generation unit, the system can directly convert natural language descriptions or multimodal analysis results into executable commands. This significantly shortens the path from task requirement to physical action. Beyond robotics, this capability can be extended to various hardware forms such as wearable devices and intelligent manufacturing robotic arms, realizing a true ``perception–cognition–decision–execution" closed loop.
	\item Self-Learning and Sustainable Evolution
	
	Leveraging a brain-inspired memory replay mechanism, ``Wenlu" can reinforce high-value information through repeated decision-making and task execution. It can also incorporate new domain-specific or private data at relatively low cost. This allows the system to continuously evolve through iterative enhancement without the need for frequent full retraining of the large model.
	\item Balance Between Generalization and Specialization
	
	Through its ``general-to-specialized" model integration strategy, ``Wenlu" retains the broad coverage of general knowledge offered by large models while demonstrating professional-level performance in vertical domains. This dual capability is particularly valuable in complex environments such as enterprise operations, healthcare, and autonomous driving, where intelligent decision-making requires both high precision and domain adaptability.
\end{enumerate}

\subsection{Ecosystem and Expansion Potential}\label{sec:5.3}
\begin{enumerate}
	\item Cross-Industry, Multi-Domain Deployment
	
	Leveraging its architecture for multimodal integration and private data management, ``Wenlu" can provide tailored solutions for diverse sectors including healthcare, finance, manufacturing, and transportation. The professional knowledge and experience accumulated within each industry will, in turn, feed back into the system, fostering a virtuous cycle of data-driven and model-driven co-evolution.
	\item Compatibility with Existing Large Model Ecosystems
	
	``Wenlu" is compatible with mainstream open-source general-purpose models (such as DeepSeek and the GPT family), enabling both migration to public platforms and customized development. This flexibility allows it to meet broader application needs while continuously integrating cutting-edge research from the AI community.
	\item Continual Evolution Through Learning
	
	The system’s brain-inspired memory replay mechanism offers diverse opportunities for future experimentation and enhancement-such as incorporating causal inference models or reinforcement learning frameworks-to further improve its adaptability to complex scenarios and long-term dynamic environments.
\end{enumerate}

In summary, the ``Wenlu" embodied intelligence system breaks through traditional boundaries between multimodal integration, privacy protection, and hardware-level decision-making. On the application front, it demonstrates strong scalability and portability. Through its forward-looking architecture and neuro-inspired memory management, the system provides an innovative approach and technical pathway for the deep industry implementation of AI and its continual self-evolution.

\section{Typical Applications and Extensions}\label{sec:6}

Leveraging its comprehensive advantages in multimodal cognition, private data protection, and hardware-level code generation, the ``Wenlu" embodied intelligence system delivers significant application value and development potential across a wide range of industries and scenarios. The following outlines its typical applications and possible directions for expansion from multiple perspectives.

\subsection{Intelligent Decision Support}\label{sec:6.1}

\subsubsection{Enterprise Management and Business Strategy}\label{sec:6.1.1}

The ``Wenlu" system can semantically analyze massive volumes of textual reports, market data, and expert documentation. By integrating this with real-time financial indicators and environmental sensing data, it supports senior executives in making scientifically informed decisions.

The private knowledge module ensures that confidential corporate documents and financial data are strictly protected during use, effectively reducing the risk of data leakage.

The multimodal decision-making capability allows unified input of diverse data formats such as text, tables, and images, enabling more comprehensive and accurate business strategy formulation.

\subsubsection{Complex Forecasting and Early Warning Analysis}\label{sec:6.1.2}

For scenarios requiring cross-domain, multi-stage data analysis-such as financial markets or supply chain management-``Wenlu" can deeply integrate text-based news, historical trading records, and sensor-based monitoring indicators to generate risk alerts or trend forecasting reports.

The brain-inspired memory replay mechanism enables the system to tag and reinforce high-risk events or abnormal fluctuations that have recently occurred, continuously enhancing its ability to respond to similar incidents in the future.

\subsection{Robotics and Hardware Control}\label{sec:6.2}

\subsubsection{Service Robots and Smart Homes}\label{sec:6.2.1}

By combining multimodal decision-making with automatic hardware code generation, the ``Wenlu" system provides an end-to-end solution for service-oriented robots-from voice understanding to action execution. For instance, a home robot can directly perform tasks such as cleaning, transporting objects, or remote monitoring upon receiving a voice command.

The private knowledge management module protects users’ household privacy data while allowing the robot, when authorized, to access restricted information (e.g., home security configurations) and perform corresponding actions.

\subsubsection{Industrial Manufacturing and Unmanned Production Lines}\label{sec:6.2.2}

In the domain of intelligent manufacturing, ``Wenlu" can analyze sensor data, surveillance images, and production logs to assess the operational status of production lines in real time. It can then automatically generate control scripts for industrial robots to carry out flexible manufacturing tasks.

In the event of anomalies, the multimodal decision module can collaborate with private knowledge (such as proprietary vendor technologies) to locate faults, while the hardware generation unit dispatches temporary control commands to mitigate risks-thus enabling a highly reliable, low-latency production system.

\subsection{Multimodal Data Processing Across Industries}\label{sec:6.3}

\subsubsection{Medical Imaging Diagnosis}\label{sec:6.3.1}

In medical scenarios, the system can integrate patient textual records, imaging results (such as X-rays, CT scans, MRIs), and physiological signals to provide more accurate diagnostic suggestions or treatment recommendations.

The private data unit ensures that sensitive patient information and internal hospital records remain confidential. Meanwhile, the general model fusion unit deeply couples publicly available medical knowledge with proprietary institutional databases, forming a personalized diagnostic assistant.

\subsubsection{Autonomous Driving and Intelligent Transportation}\label{sec:6.3.2}

For autonomous vehicles, the system can comprehensively analyze visual imagery, radar/LiDAR sensor data, voice navigation commands, and traffic flow information to generate optimal driving strategies in real time.

When confidential maps or safety protocols are involved, the private knowledge unit manages these via encryption and retrieves the necessary data for critical decisions such as route planning and traffic prediction. The automatically generated execution scripts are then directly applied to vehicle controllers, enabling highly autonomous vehicle operations.

\subsection{Next-Generation Human-Computer Interaction Systems}\label{sec:6.4}

\subsubsection{Intelligent Customer Service and Expert Systems}\label{sec:6.4.1}

Enterprises or public institutions can deploy the ``Wenlu" system in customer service domains to provide consultations and services that integrate multimodal inputs such as voice, text, and images, significantly enhancing user interaction experience.

Through the ``general-specialized" combined knowledge base fusion approach, the customer service system can handle not only general inquiries but also invoke deep industry knowledge to offer authoritative answers in specialized scenarios such as legal consultation and insurance claims.

When user private data is involved (e.g., ID cards, policy information), the private decision unit strictly controls access, outputting customized responses after security review.

\subsubsection{Virtual Assistants and Wearable Devices}\label{sec:6.4.2}

The system can interface with AR/VR devices, smartwatches, smart glasses, etc., analyzing real-time camera footage or sensor data, generating personalized suggestions via the language model, and delivering feedback to users in natural language.

In healthcare scenarios, virtual assistants can access sensor data such as heart rate, blood pressure, and step counts, then combine this with the general model’s health knowledge base to provide exercise and diet recommendations while strictly complying with user privacy protection requirements.

\subsection{General-Purpose AI Large Model Platform}\label{sec:6.5}

\subsubsection{Integration of Open-Source Models and Industry Data}\label{sec:6.5.1}

``Wenlu" can deeply integrate with various open-source large models by performing targeted fine-tuning or incremental learning on general models, efficiently coupling industry data, private information, and public corpora.

Relying on the foundational language understanding and generation capabilities provided by general models, supplemented by proprietary industry knowledge bases and private data management, the system achieves a next-generation intelligent core with stronger specificity and scalability.

\subsubsection{Self-Learning Ecosystem}\label{sec:6.5.2}

The system accumulates extensive multimodal task data through repeated interactions and continuous use, forming a self-learning cycle combined with a brain-inspired memory replay mechanism.

In the long run, ``Wenlu" will continuously enhance its mastery over key tasks and algorithmic strategies across various industries, advancing from a general large model platform toward a broader general artificial intelligence ecosystem.

\subsection{Future Expansion Directions}\label{sec:6.6}
\begin{enumerate}
	\item Deeper Causal Reasoning and Explainable Machine Learning
	
	By integrating causal reasoning frameworks with reinforcement learning concepts, further enhance the system’s explainability and robustness in complex decision-making and uncertain scenarios.	
	\item Cross-Modal Generation and Reverse Engineering
	
	Beyond code generation, explore multimodal generation capabilities such as text-to-image and text-to-3D motion simulation to support fields like design, scientific research, and the arts.	
	\item Multi-Level Privacy and Compliance Strategies
	
	For stricter or more detailed data compliance scenarios (e.g., medical privacy laws, financial regulations), implement finer-grained access control and auditing functions within the private knowledge decision unit.
\end{enumerate}

In summary, the ``Wenlu" Embodied Brain system demonstrates significant advantages in multimodal fusion, efficient decision-making, and private data protection across multiple industry applications. In fields such as healthcare, finance, industrial manufacturing, and next-generation human-computer interaction, the system provides end-to-end intelligent services. With its continuous learning mechanisms, it delivers an AI experience characterized by both breadth and depth. As large models achieve further breakthroughs in multimodal understanding, ``Wenlu" will continue to possess broad innovation and expansion potential, laying a solid foundation for the development of next-generation general intelligent platforms.

\section{Specific Implementation Methods and Workflow}\label{sec:7}

To maximize the effectiveness of the ``Wenlu" Embodied Brain system in practical applications, it is necessary to design and implement a systematic process spanning from initialization and deployment to daily operation. This section focuses on the coordination and information flow among the system’s main modules, explaining the typical workflow and detailing key implementation points.

\subsection{Initialization and Model Loading}\label{sec:7.1}
\begin{enumerate}
	\item General Model Deployment
	\begin{itemize}
		\item Deploy open-source or commercial large-scale language models (e.g., DeepSeek) in cloud or local server environments, with appropriate computing resources provisioned (GPU/TPU).
		\item Perform necessary fine-tuning or incremental training based on the target industry requirements to improve the model’s accuracy in question-answering and reasoning within specific domain scenarios.
	\end{itemize}
	\item Industry Knowledge Base and Private Data Management Module Preparation
	\begin{itemize}
		\item Import or build industry-specific knowledge bases (such as medical, manufacturing, finance) to provide robust data support for multimodal decision-making and reasoning.
		\item Configure the private knowledge decision unit to encrypt, label, and index materials that may contain sensitive information (texts, tables, images, confidential documents, etc.).
		\item Establish access control policies, including role-based permissions and desensitization output mechanisms, ensuring private data is stored securely and separately from public training corpora.
	\end{itemize}
	\item Multimodal Input Channel Construction
	\begin{itemize}
		\item Interface with image capture devices, voice acquisition, sensor data reading, etc., to achieve real-time connection or periodic data retrieval from external devices/databases.
		\item For scenarios requiring physical interaction (e.g., robots, wearable devices), initialize communication protocols or adaptation layers to ensure smooth command sending and receiving post-deployment.
	\end{itemize}
\end{enumerate}

\subsection{Encrypted Access and Tagging of Private Data}\label{sec:7.2}
\begin{enumerate}
	\item Private Data Upload and Identification
	\begin{itemize}
		\item Users or system administrators upload private files via the backend (e.g., internal corporate documents, medical records, confidential technical plans), which are automatically encrypted by the system and assigned a unique identifier.
		\item Through text analysis or predefined metadata, the system performs an initial semantic parsing of the file contents, generating tags for sensitivity level, associated topics, and access permissions.
	\end{itemize}
	\item Permission Management and Indexing
	\begin{itemize}
		\item Based on Role-Based Access Control (RBAC) strategies, each private data item is assigned an access level (e.g., visible only to administrators or specific business units).
		\item A private index table is constructed within the system, embedding document content as vectors or keywords and linking them to the general model for subsequent use in question-answering or inference.
	\end{itemize}	
	\item Encrypted Storage and Sandbox Isolation
	\begin{itemize}
		\item A secure sandbox mechanism is employed to store private data separately from public corpora, and encryption algorithms (symmetric or asymmetric) are used to protect the files.
		\item When external applications initiate requests involving private information, they must first pass an authorization and verification process before they can read or write sensitive content.
	\end{itemize}
\end{enumerate}

\subsection{Multimodal Data Fusion and Preprocessing}\label{sec:7.3}
\begin{enumerate}
	\item Feature Extraction Module
	\begin{itemize}
		\item For image data, convolutional neural networks or other vision models are used for object detection, classification, or feature embedding. For audio data, acoustic feature extraction or speech recognition models are employed. Sensor data is processed through filtering, normalization, and other techniques.
		\item Raw information from different modalities is transformed into unified or alignable vector representations and stored in an intermediate buffer.
	\end{itemize}	
	\item Semantic Fusion and Contextual Encoding
	\begin{itemize}
		\item Leveraging the multimodal decision unit within the ``Wenlu" system, feature vectors from images, speech, text, and other modalities are integrated into a shared contextual encoding model. Attention mechanisms are applied to learn semantic associations across modalities.
		\item When needed, the system can dynamically access the domain-specific knowledge base to identify professional knowledge points embedded in features (e.g., lesion locations in medical images or fault points in industrial inspection data).
	\end{itemize}	
	\item Privacy Detection and Filtering
	\begin{itemize}
		\item If any multimodal data contains content that may be mapped to private indices (e.g., internal identifiers, confidential parameters), a privacy identification check is first performed. If triggered, the system coordinates with the private knowledge module to verify access rights before determining the next steps.
		\item Content without sensitive markings is directly passed to the general model for deep semantic parsing and reasoning.
	\end{itemize}
\end{enumerate}

\subsection{Question Answering and Decision-Making Inference Workflow}\label{sec:7.4}
\begin{enumerate}
	\item User Requests / Task Inputs
	\begin{itemize}
		\item Users may initiate queries or decision-making requests via natural language, multimodal inputs (text-image combinations), or voice commands. Tasks can also be triggered by external events, such as industrial equipment failure alerts or urgent medical diagnostic needs.
		\item If the task involves references to private data or requires high-sensitivity decision-making, the system will immediately invoke the private knowledge module to verify access permissions.
	\end{itemize}	
	\item Inference by the General Model Integration Module
	\begin{itemize}
		\item The system feeds the user request, multimodal inputs, and domain-specific knowledge (including private data, if access is authorized) into the general-purpose language model for inference.
		\item A brain-inspired memory tagging mechanism is introduced to preliminarily annotate key knowledge points or reasoning paths involved in the current inference process.
	\end{itemize}	
	\item Decision Output or Question Answering Result
	\begin{itemize}
		\item The general model integration module produces an initial decision outcome or response, while the multimodal decision-making module annotates and interprets visual outputs (e.g., diagnostic maps).
		\item If private or sensitive information is involved, the system conducts a desensitization check prior to output, ensuring that the generated text or images do not exceed the user's permission scope.
	\end{itemize}
\end{enumerate}

\subsection{Embodied Code Generation and Execution}\label{sec:7.5}
\begin{enumerate}
	\item Triggering Hardware Control Requests
	\begin{itemize}
		\item Once the system’s decision outcome includes requirements for robotic actions, wearable device scheduling, or other hardware control tasks, the hardware control generation module is immediately activated.
		\item For example, a user might input, ``Please control the robotic arm to grasp the red object and place it in area X," or the multimodal detection module may determine that ``a specific valve needs to be shut off to prevent further escalation of a malfunction."
	\end{itemize}	
	\item Automated Script / Command Generation
	\begin{itemize}
		\item Based on ROS2 or the target hardware’s API specifications, the hardware control generation module translates natural language inputs or multimodal decision results into executable control scripts (e.g., in Python, C++, etc.).
		\item For different models or brands of robotic systems, a low-level adaptation layer translates the standardized intermediate instructions into device-specific command formats compatible with the appropriate drivers.
	\end{itemize}	
	\item Execution and State Feedback
	\begin{itemize}
		\item The generated scripts or commands are dispatched in real time to the designated hardware. If the device supports feedback (e.g., via sensors or status logs), the returned data is automatically routed into the multimodal analysis module for secondary evaluation.
		\item If obstacles or environmental changes arise during execution, the system can re-infer and quickly generate an updated control strategy, achieving a dynamic, adaptive closed-loop process.
	\end{itemize}
\end{enumerate}

\subsection{Neuro-Inspired Memory Tagging and Replay}\label{sec:7.6}
\begin{enumerate}
	\item Key Decision Point Tagging
	\begin{itemize}
		\item After completing a full cycle of interaction, reasoning, or hardware operation, the system attaches memory annotations to critical information and reasoning steps, recording their weight distributions in multimodal fusion and knowledge base retrieval.
		\item If private data was accessed during the task, the access path is securely recorded to facilitate future auditing and policy optimization.
	\end{itemize}	
	\item Offline Replay and Reinforcement
	\begin{itemize}
		\item During system idle periods or in offline batch processing, the system replays a set of previously annotated critical decisions. Reinforcement learning or parameter tuning techniques are used to improve performance and accuracy on similar tasks.
		\item This continual learning process does not require retraining the entire foundation model but only performs lightweight fine-tuning on internal strategy parameters or fusion modules.
	\end{itemize}	
	\item Index and Policy Updates
	\begin{itemize}
		\item After replay, the system writes the improved strategies into the knowledge index or private data policy modules, allowing future tasks to directly benefit.
		\item This forms a positive feedback loop: multi-turn interaction $\rightarrow$ memory tagging $\rightarrow$ offline reinforcement $\rightarrow$ performance enhancement, driving self-evolution of the system.
	\end{itemize}
\end{enumerate}

\subsection{System Maintenance and Expansion}\label{sec:7.7}
\begin{enumerate}
	\item Incorporating New Industry Knowledge Bases
	\begin{itemize}
		\item When expanding to new industries (e.g., agriculture, construction, energy), relevant knowledge bases can be imported via incremental learning, with appropriate model plugins added to the multimodal feature extraction stage.
		\item The general language model only requires lightweight parameter updates, minimizing redundant training overhead.
	\end{itemize}
	\item Updating Privacy Policies and Compliance Protocols
	\begin{itemize}
		\item In response to evolving legal regulations or corporate compliance needs, the private knowledge unit's access rules, tagging schemes, and anonymization algorithms can be flexibly updated.
		\item Encryption algorithms and security protocols are regularly reviewed to ensure the system remains up-to-date with privacy and compliance standards.
	\end{itemize}	
	\item Hardware Interface and Adaptation Layer Maintenance
	\begin{itemize}
		\item As hardware is upgraded or new platforms are introduced, only the adaptation layer scripts and API mappings in the hardware control generation module need to be updated-without affecting the higher-level decision logic.
		\item During long-term operation, data from various hardware failures or anomalies can be collected to further improve the robustness of code generation.
	\end{itemize}
\end{enumerate}

\subsection{Operational Example: Multimodal Diagnosis and Robotic Execution}\label{sec:7.8}
\begin{enumerate}
	\item Scenario Description
	
	Industrial scenario: A production line triggers a fault alert. The system receives data streams from sensors (e.g., temperature, pressure), images from surveillance cameras, and textual descriptions from operators.
	\item Multimodal Fusion and Privacy Determination
	
	The multimodal decision unit performs feature extraction to identify the fault location. If proprietary component information is involved, the private knowledge unit performs secure access validation.	
	\item General Model Reasoning and Output
	
	Combining the industry knowledge base and general model capabilities, the system identifies the most probable fault cause and recommends component replacement or parameter adjustments.
	\item Automated Robot Maintenance Command Generation
	
	The hardware control module translates the fault-handling strategy into robotic movement, grasping, and inspection scripts. The robot executes the instructions and sends feedback data for secondary verification.	
	\item Offline Replay and Reinforcement Learning
	
	During idle periods, the system replays the decision path, logs fault characteristics, repair methods, and execution efficiency, contributing to enhanced future responses.
\end{enumerate}

Through the above end-to-end workflow, the ``Wenlu" Embodied Brain System continuously assimilates multimodal data, industry expertise, and private user information to enable self-learning and capability iteration with minimal retraining cost. At the same time, its end-to-end hardware execution capability provides a closed loop from cognition to action, significantly shortening the cycle from intelligent decision-making to real-world deployment, while ensuring high reliability and security in critical application scenarios.

~\\

\par\noindent
\parbox[t]{\linewidth}{
	\noindent\parpic{\includegraphics[width=1in,height=1.25in,clip,keepaspectratio]{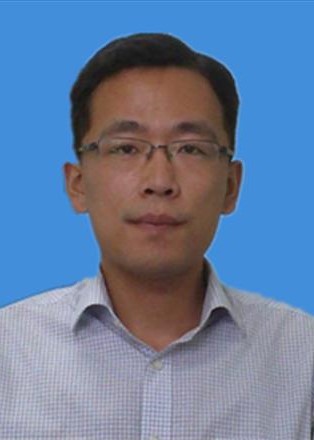}}
	\noindent {\bf Liang Geng}\
	received the master's degree from Hebei University of Technology, Tianjin, China, in 2012. He is currently pursuing a Ph.D. degree with the School of Artificial Intelligence, Beijing University of Posts and Telecommunications, Beijing, China. He is also an Assistant Research Fellow at the College of Mechanical and Electrical Engineering, Shijiazhuang University, as well as at the Key Laboratory of Agricultural Robotics Intelligent Perception in Shijiazhuang, China.}
\vspace{0.5\baselineskip}

%Bibliography
%\bibliographystyle{unsrt}  
%\bibliography{references}  

\end{document}